# Information Modeling for a Dynamic Representation of an Emergency Situation

Fahem Kebair and Frédéric Serin

*Abstract*—In this paper we propose an approach to build a decision support system that can help emergency planners and responders to detect and manage emergency situations. The internal mechanism of the system is independent from the treated application. Therefore, we think the system may be used or adapted easily to different case studies. We focus here on a first step in the decision-support process which concerns the modeling of information issued from the perceived environment and their representation dynamically using a multiagent system. This modeling was applied on the RoboCupRescue Simulation System. An implementation and some results are presented here.

## I. Introduction

Managing an emergency situation usually includes a lack of knowledge, unsure information and a delicate coordination. It is therefore difficult to actors to make good decisions in time and to coordinate efficiently their efforts, since they do not have enough knowledge about the situation or they do not have timely information they need.

The emergency response is one of the greatest challenges that arise to the society currently. One approach to address this challenge is to develop decision support systems (DSS) that may help improve emergency planners and responders awareness and their decision-making abilities. Moreover the system must anticipate the risk of calamitous events or the evolution of a current crisis. This makes planners warned and prepared permanently to future events. Consequently, they can produce robust plans towards both short-term and long-term goals.

A number of research efforts have been targeted at this topic with the aim to create modeling techniques and tools for the emergency management, some of them are described in [6]. The research presented in this paper is situated in the heart of this problematic with the volition to build a generic DSS. We mean by genericity that the system is not addressed to a particular application or a specific subject of study. Hence, we deal currently with several case studies in order to test the system. However, the work done so far concerns essentially a part of the DSS which intends to represent the information about the current situation. A first prototype dedicated to the game of Risk was developed [12]. Then, we started working on a new prototype version [8] addressed to the RoboCupRescue simulation system (RCRSS) [13]−[9] and on which we focus in this paper and finally a study in the e-learning domain is ongoing [2].

Moreover, flexibility remains crucial to the success of planning and response operations [11]−[14]. Traditional systems are known to be less or not adaptive. This hypothesis incited us to endow the system with adaptivity and flexibility abilities. We are thinking then that the multiagent approach is the most suitable technology for use, since intelligent agents and multiagent systems (MAS) technologies are considered to be a promising method to construct the scalable, robust, reusable high quality software system.

The system proposed here has an internal multiagent layered kernel. Agents used here are as defined by Jennings in [7]. This architecture must respond in a first step to one of the aspects of our approach which is based on the construction of an environment perception model. A reification of the perceived environment is therefore necessary to define the observations that can be extracted from the observed situation. Information issued from these observations are embedded into agents which aim to represent dynamically the evolution of the current situation. Emergent agents are then characterised and analysed to form clusters. The comparison of each cluster with other situations previously defined will provide finally outcomes to planners.

This paper is structured as follows; first, the proposed architecture of the system is presented. Then, a modeling of an emergency situation and an approach to formalise information that describe are illustrated. Next, a description of the representation layer is provided. Afterwards, an implementation and tests are described. Finally, a conclusion and perspectives are mentioned.

## II. Agent-based decision support system

The role of the DSS is quite wide. In general, the purpose is "to improve the decision making ability of managers (and operating personnel) by allowing more or better decisions within the constraints of cognitive, time, and economic limits" [5]. More specifically, the major characteristics of a DSS are:
- DSS incorporate both and models.
- They are designed to assist managers in semi-structured or unstructured decision-making process;
- DSS support, rather than replace, managerial judgment.

Fahem Kebair is a PhD candidate in computer science in LITIS (Laboratoire d'Informatique de Traitement de l'Information et des Systèmes) in Le Havre University (address : 25 rue Philippe Lebon, 76058, Le Havre Cedex, France, e-mail: fahem.kebair@univ-lehavre.fr).

Frédéric Serin is a professor assistant in computer science in LITIS in Le Havre University (telephone: +33-02-327-44384, e-mail : frederic.serin@univ-lehavre.fr).

- They are aimed at improving the effectiveness rather than efficiency of decisions.

According to Turban and Aronson [15], the central purpose of a DSS is to support and improve decision making and to be successful, such a system needs to be adaptive, easy to use, robust and complete on important issues [10]. In the context of the emergency response, where the environment is dynamic and uncertain, these characteristics are essential for the system to support planners and responders to manage the emergency.

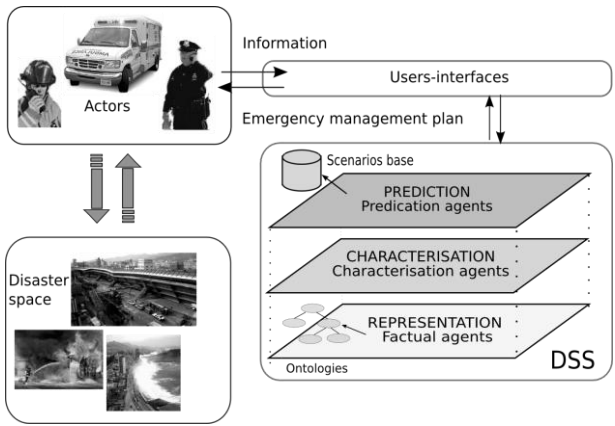

Fig. 1. DSS architecture

Fig. 1 shows the architecture of the DSS. The system is feed permanently by information describing the environment state. To deal with these data, the system needs some knowledge about the environment as the ontologies of the domain. The outcomes provided by the system include an evaluation of the situation and an emergency management plan to manage it. The system has a multiagent layered kernel whereof mechanism is the following;

The first step is to deal with information coming from the environment thanks to factual agents. These agents form a representation layer that has as essential role to represent dynamically the current situation. Each factual agent aims to reflect a partial part of the observed situation. Its purpose is to reach a predominant place in the factual MAS by fighting some agents and helping some others.

The second step is to analyse the emergent organisation of factual agents by creating agents in a characterisation layer. This MAS is generated using dynamic clustering techniques. Each cluster of factual agents leads to the creation of a characterisation agent (or clustering agent). This agent represents a class of facts which emerge by similarity between diverse factual agents and dissimilarity between characterisation agents. This clustering is not necessarily supervised since one of our objectives is to detect a risk not inevitably referenced as such; it is dynamic because the observed entering facts modify permanently the structure of the factual MAS and dysfunctions may evidently appear during the observation or, on the contrary disappear.

The third step is to identify scenarios that will be carried by prediction agents. These agents constitute a prediction layer and aim to find clusters, in the characterisation layer, enough close to inform the decision-makers about the current situation and its probable evolution, verily to generate a warning in the case of detecting a risk of crisis. This mechanism is managed by a Case-Based Reasoning (CBR) and is studied "manually" by the expert of the case study. The CBR must provide the possible consequences of a given scenario and the solution that match to this particular case of the situation.

III. MODELING AND DYNAMIC REPRESENTATION OF INFORMATION

A. Observed Environment Reification

A reification of the observed environment using the object paradigm is needful to format the observations about this environment in factual semantic features (FSF). Each observation is dual and provides a description, at the same time, of concrete entities and a dynamics of the environment. From diverse concrete cases we propose decomposition in six classes that may be qualified as abstract or generic (Fig. 2). These classes belong to two families or-in other words-inherit two super classes.

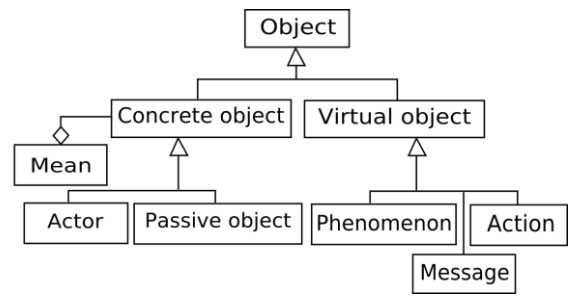

Fig. 2. Observations object-modeling

The first one is constituted by individual concrete objects or exceptionable composite; it gathers the three classes that are passive entities or objects of the environment, actors which are active entities and means which are a collection of various concrete objects. These classes describe therefore direct and concrete observations concerning passive and active entities; means are introduced to characterise a set of observable objects of which we could specify related behaviours such as the changes of correlated states, concentrated behaviours or physical grouping that infer to emergent and specific actions of this gathering.

The second family categorises the virtual objects which are the different forms of activities; it gathers the three other classes which are phenomena, actions and messages. The elements of this family are deduced from indirect observations and are formalised basing-on the "memento" design pattern of Gamma [4]. They are differentiated by their interconnections with concrete objects. Actions are related to actors or means, filled with goals, and take action on the states of other concrete objects. Phenomena have no goals for the least intelligible neither identifiable actors that trigger them. Messages are very close to actions and are distinguished by the lack of change in states without intermediate pretreatment verily without any notable change.

B. Application on RoboCupRescue

We choose the RCRSS as a test bed because it is well suited for the emergency response and as a game, we can have a perfect knowledge about its environment. RCRSS is an agent-based simulator which intends to reenact the rescue mission problem in real world. An earthquake scenario is reproduced in the RoboCupRescue (RCR) environment including various kinds of incidents as the traffic after earthquake, buried civilians, road blockage and fire accidents. A set of heterogeneous agents coexist in the disaster space, each with a specific goal and a particular role.

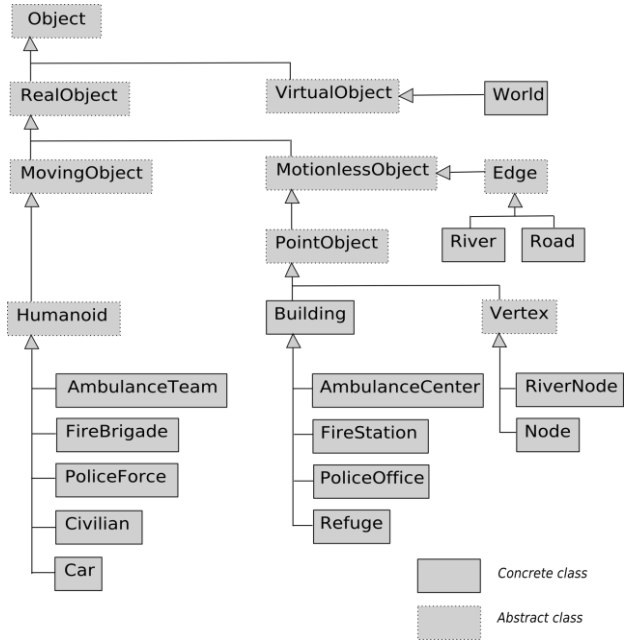

Fig. 3. Class hierarchy of the RCR objects in the disaster space

Fig. 3 shows an object-modeling describing the RCR environment. From this model we were able to identify the concrete observable entities that we have defined in the previous section. Actor objects are modelled as Humanoid objects and Passive objects are modelled as Motionless objects. Virtual objects in the RCR hierarchy class have a completely different meaning; they represent the knowledge about the disaster world acquired by each RCR agent.

### C. Information Formalisation

Formatting the reified observations according to the environment object-modeling (paragraph A) and the RCR class hierarchy (paragraph B) introduces for our system the fundamental notion of FSFs. The given noun to this message content brings an account to our approach: we stress observed and punctual elements that are facts namely observed elements directly and beforehand deduced that we estimate to be bearer of meaning for the grasp, the appropriation and the analysis of the environment. FSFs are therefore initial elements that permit to detect risks. They are the messages content referring to the FIPA denomination [3].

Each FSF is composed of an object that describes. This object is associated to its class issued from the model specialisation presented above. To each object, the FSF associates qualifiers and their related values at the time of the observation. These qualifiers refer to the objects states and are incorporated into ontologies of the studied domain. An example of an FSF related to a fire phenomenon is: (fire#14, fieriness, 1, inDangerNeighbours, 3, burningNeighbours, 2, localisation, 20|25, time, 7)

This FSF means a fire is ignited in building number 14 with intensity equals to 1, has 2 burning fires and can spread to 3 neighbour buildings. The burning building has the following coordinates 20|25, and the fire was perceived in the 7$^{th}$ cycle[1] of the simulation.

In order to make emerge the significant facts of the current situation, we have introduced the proximity notion between FSFs. For every application, the proximity measure returns a value in [-1 .. 1]. The more a proximity value between two FSFs is close to 1 the stronger the semantic connexion is, and vice versa. We distinguish three types of proximities: semantic proximity $P_s$, temporal proximity $P_t$ and spatial proximity $P_e$. The total proximity $P$ is the product of the three proximities. $P_s$ is computed basing on the specific ontology. $P_t$ and $P_e$ are distances respectively in time and space. They are defined in an interval of [0, 1] and are computed using these formulas:

$$P_t = (4 \exp(-0.2\Delta t)) / (1 + \exp(-0.2\Delta t))^2$$
$$P_t = (4 \exp(-0.08\Delta e)) / (1 + \exp(-0.08\Delta t))^2$$

Where $\Delta t$ is the difference of time and $\Delta e$ is the Euclidean distance between the two objects.

### D. Representation MAS

The acquisition of the FSFs sent to the system is insured by the representation layer which is composed of factual agents. The kind of this treatment is directly resulting from the proximity value between the diffused observation and the information already contained in each agent. In order to not lose any information, a generative agent creates a new factual agent that will incorporate the original information, if the approbation by existent agents is estimated insufficient (effect of semantic proximities threshold).

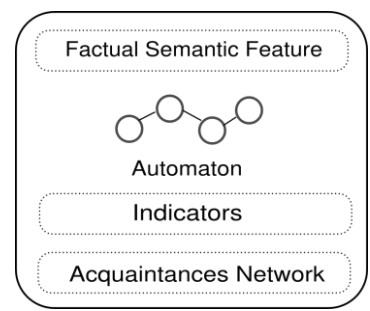

Fig. 4. Internal structure of a factual agent

Each factual agent has an internal automaton which is an augmented transition network (ATN) and that insures the proactiveness of the agent. The ATN is composed by a set of states, to which generic and specific actions are associated, linked by transitions carrying specific conditions on the agent properties.

A generic acquaintances network is used by each factual agent to interact with the other agents. This network is a dynamic memory of agents whereof proximity values with the current agent is different from 0.

---
[1] One cycle in the simulation equals one second

Basing-upon the object model presented in section III, we defined two categories of factual agents in RCR: factual agents related to RCR agents (actors of the environment) and factual agents related to phenomena. Each kind of factual agent has its own behaviour and its own purpose in order to reflect as close as possible the real world. This characteristic has thereby an impact on the definition of the internal automata and the compute of the indicators of the factual agents.

Each factual agent has two indicators to reflect its dynamics. These indicators provide a synthetic view of the salient facts of the situation. In the case of the RCRSS we have defined two indicators associated to a factual agent:

- Action indicator (AI): it represents the position and the strength of a factual agent inside the representation MAS. For factual agents related to RCR agents, AI means the potential of an RCR agent and its efficiency in solving a problem. For factual agents managing phenomena, AI means the degree of damage and hazard that could represent this phenomenon.
- Plausibility indicator (PI): For factual agents related to RCR agents, PI means the ability of an RCR agent to discover new problems in the disaster space. For phenomena factual agents, PI means the solving probability and the worsening impediment of a phenomenon.

As follows a way to compute AI and PI for a fire factual agent:

$$AI = AI + proximity (FSF1, FSF2)$$
$$PI = 10 \exp(-0.05x) + proximity (FSF1, FSF2)$$

Where x = (burningNeighbours + fieryness + lifeTime) – 5×nbFireBrigades
lifeTime: time since the creation;
nbFireBrigades: number of fire brigades around the fire.

## IV. IMPLEMENTATION AND TESTS

We choose JADE [1] platform to implement the representation MAS because it is FIPA compliant, but also because jade agents are suitable to implement factual agents.

The implementation and the tests carried out so far include a part of the ontology dealing with fires. Therefore, we treat here fires factual agents and fire brigade factual agents. Fig. 5 shows two internal ATNs. The upper one belongs to a fire factual agent and the lower one belongs to a fire brigade factual agent.

Each agent has an automaton in four states. The first state is a creation state in which the agent is created and enters in activities. State 2 and 3 are the main states of the ATN, wherein the agents are very active. In state 4, the factual agent is dead and ceases its activities. Both fire and fire brigade factual agents change state when they satisfy transitions conditions that are defined as indicators thresholds.

A fire factual agent is in state 3 when it has high values of AI and PI. This means the fire is important or there are fire brigades who are close to it and that can extinguish it. When the fire is increasingly irrelevant, the related factual agent regresses in state 2, due to the decrease of its AI value.

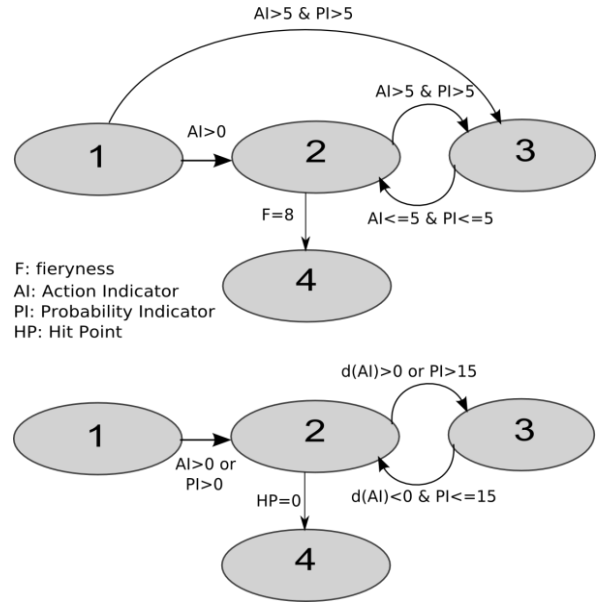

Fig. 5. ATNs of fire and fire brigade factual agents. d(AI) is the difference between two successive values of AI.

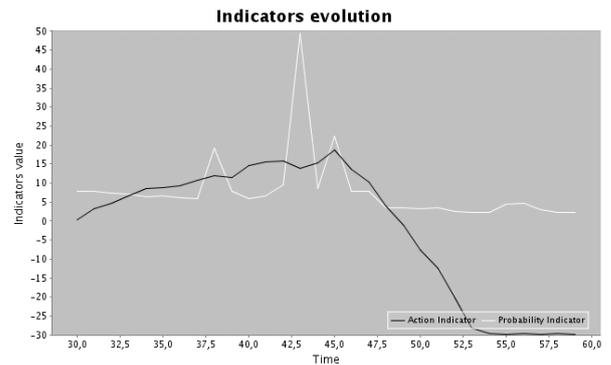

Fig. 6. Indicators evolution of a fire factual agent.

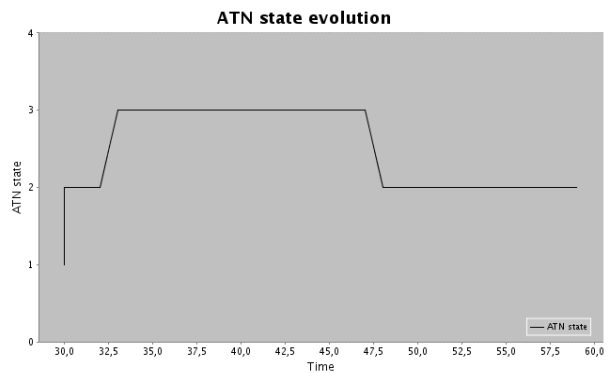

Fig. 7. ATN state evolution of a fire factual agent

Fig. 6 and Fig. 7 show respectively the evolution of a fire factual agent indicators and its state inside the ATN over time. The fire here is created at the 30th cycle of the simulation. The fire is important at its beginning, because on the one hand it can spread to its neighbourhood and on the other hand it is easier to extinguish. Thereby, the fire moves rapidly to state 3 in the 32th cycle and still in it until the 48th cycle where the fire brigades extinguished it. The three peaks of PI noted between the 38th and the 46th cycles, explain the presence of fire brigades in the fire location. The height of these peaks differs according to the number of fire brigades and their actions, as in cycle 43th when they start extinguishing the fire.

Fig. 8 and Fig. 9 show respectively the evolution of a fire brigade factual agent indicators and its state inside the ATN over time. We notice a greater oscillation of the chart for this category of factual agents then the one of the fire factual agent and this due to the permanent move of the fire brigades, unlike fires which are located on the buildings, which are immobile.

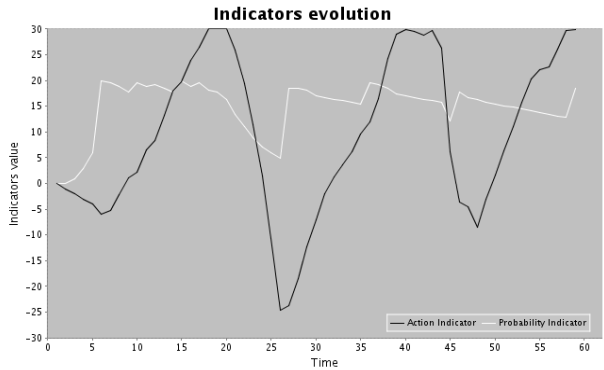

Fig. 8. Indicators evolution of a fire brigade factual agent

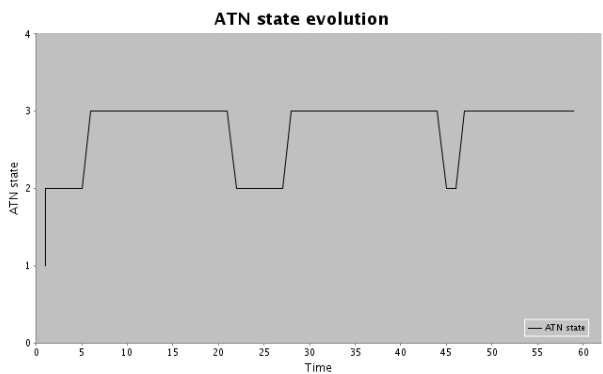

Fig. 9. ATN state evolution of a fire brigade factual agent

A fire brigade factual agent is in state 2 when it searches new fires or when it tries to join other groups. In this state the factual agent has low values of AI and PI. However, when the fire brigade discovers new fires, or when he is extinguishing a fire, his factual agent transits to states 3 due to the increase of its AI and PI values. We notice also that every variance of PI is followed by a variance of AI. This justifies the change in the behaviour of the fire brigade after having discovered a new fire.

Fig. 10 shows the number of activities of all the factual agents during each cycle. Activities include the change of state in the ATN and the change of indicators values. We notice here an increase of activities from the 5$^{th}$ cycle, when the fire brigades discover the first fires. The number of activities then grows during the simulation, due to fire spread to which the fire brigades will respond. The chart presents also an oscillation that explains the behaviour of the fire brigades that work generally by group to be able to fight fires. We have therefore a relatively high number of activities when the fire brigades discover new fires of when they are fighting fires. On the contrary, when the fire brigades are inactive or when they are exploring new areas of the city to find fires, we have fewer activities.

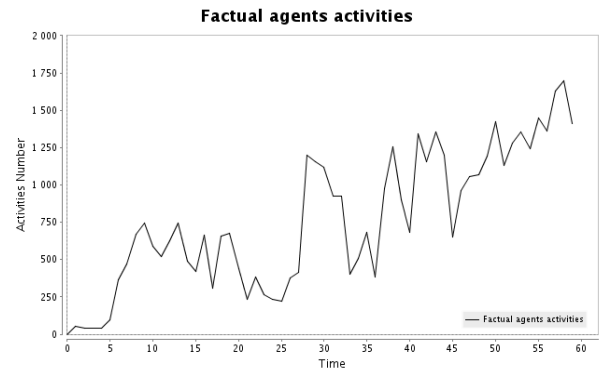

Fig. 10. Number of factual agents activities during each cycle

## V. CONCLUSION

In this paper we proposed an approach to build a system that aims to help emergency planners and responders to detect and manage emergency situations. We described here a first step of our approach which is the formalisation of information, issued from observations upon a partially and unpredictable environment, using the notion of factual semantic features.

The kernel of the system is made up of multiagent layers. We presented here the lower layer that intends to represent dynamically the semantic of the current situation and its evolution over time thanks to factual agents. We think that the main part of the kernel is generic. In our approach, it is therefore indispensable to produce various validation tests on relatively diverse applications. Among the applications on which we work currently is the RCRSS that is well suited for the emergency response. Thus, after a long study on the simulation environment, we defined and created the FSFs that allow the RCR environment description and started tests on the representation layer. Thanks to these tests we were able to validate the formalisation of the FSFs and to analyse a part of the representation MAS.

Currently, we are finishing the implementation of the representation MAS. At the same time, we are working on the middle and the upper layer of the kernel, where we intend to set up several scenarios in order to select the appropriate clustering techniques and to define the similarity measures to compare between the elaborated scenarios and the formed clusters.